\documentclass[runningheads]{llncs}

\usepackage{eccv}

\usepackage{eccvabbrv}

\usepackage{graphicx}
\usepackage{booktabs}
\usepackage{multirow}

\usepackage[accsupp]{axessibility}  

\usepackage{hyperref}

\usepackage{orcidlink}

\usepackage{todonotes}
\usepackage{xcolor}
\usepackage{graphicx}
\definecolor{darkgreen}{RGB}{0,120,0}
\definecolor{darkred}{RGB}{180,0,0}
\newcommand{\gainp}[1]{\textcolor{darkgreen}{+#1}}
\newcommand{\gainn}[1]{\textcolor{darkred}{-#1}}
\usepackage{rotating}
\usepackage{tabularx}
\usepackage{amssymb}
\usepackage{pifont}
\usepackage{adjustbox}
\newcommand{\xmark}{\ding{55}}%
\usepackage{booktabs}

\newcommand{\greencheck}{{\color{green}\checkmark}}
\newcommand{\redxmark}{{\color{red}\xmark}}

\usepackage[acronym]{glossaries}
\newacronym{ALS}{ALS}{airborne laser scanning}
\newacronym{MLS}{MLS}{mobile laser scanning}
\newacronym{LoD}{LoD}{level of detail}
\newacronym{LoDs}{LoDs}{level of details}
\newacronym{OGC}{OGC}{Open Geospatial Consortium}
\newacronym{GML}{GML}{Geography Markup Language}
\newacronym{ASAM}{ASAM}{Association for Standardization of Automation and Measuring Systems}
\newacronym{TLS}{TLS}{terrestrial laser scanning}
\newacronym{UAV}{UAV}{unmanned aerial vehicle}
\newacronym{HD}{HD}{high definition}
\newacronym{RANSAC}{RANSAC}{RANdom SAmple Consensus}
\newacronym{ROI}{ROI}{region of interest}
\newacronym{DEM}{DEM}{digital elevation model}
\newacronym{ICP}{ICP}{iterative closest point}
\newacronym{NLOS}{NLOS}{non-line-of-sight}
\newacronym{SfM}{SfM}{structure from motion}
\newacronym{FME}{FME}{Feature Manipulation Engine}
\newacronym{OSM}{OSM}{OpenStreetMap} 
\newacronym{RMSE}{RMSE}{root mean square error}
\newacronym{CPT}{CPT}{conditional probability table}
\newacronym{DST}{DST}{Dempster–Shafer theory}
\newacronym{BN}{BayNet}{Bayesian network}
\newacronym{GIS}{GIS}{Geographic Information System}
\newacronym{PPD}{PPD}{posterior probability distribution}
\newacronym{CI}{CI}{confidence interval}
\newacronym{IFC}{IFC}{Industry Foundation Classes}
\newacronym{CRS}{CRS}{coordinate reference system}
\newacronym{LoFG}{LoFG}{Level of Facade Generalization}

\begin{document}

\title{UnderOneFacade: Worldwide Facade Semantic Segmentation Benchmark Dataset} 

\titlerunning{UnderOneFacade}


\author{
Yi Wang\inst{1}$^\star$
\and
Fan Wang\inst{1}$^\star$
\and
Wanru Yang\inst{2}$^\star$
\and
Prabin Gyawali\inst{1}
\and
Ziyang Xu\inst{1}
\and
Anna Klimkowska\inst{3}
\and
Yixiong Jing\inst{2}
\and
Filip Biljecki\inst{4}
\and
Christoph Holst\inst{1}
\and
Benjamin Busam\inst{1}
\and
Brian Sheil\inst{2}
\and
Olaf Wysocki\inst{2}
}
 
\authorrunning{Y.~Wang et al.}

\institute{
Technical University of Munich, Munich, Germany \and
CV4DT, University of Cambridge, Cambridge, UK \and
University of Nottingham, Nottingham, UK \and
National University of Singapore, Singapore\\[2pt]
$^\star$ \texttt{equal contribution}
}

\maketitle
\begin{figure*}[ht]
  \centering
  \includegraphics[width=0.8\textwidth]{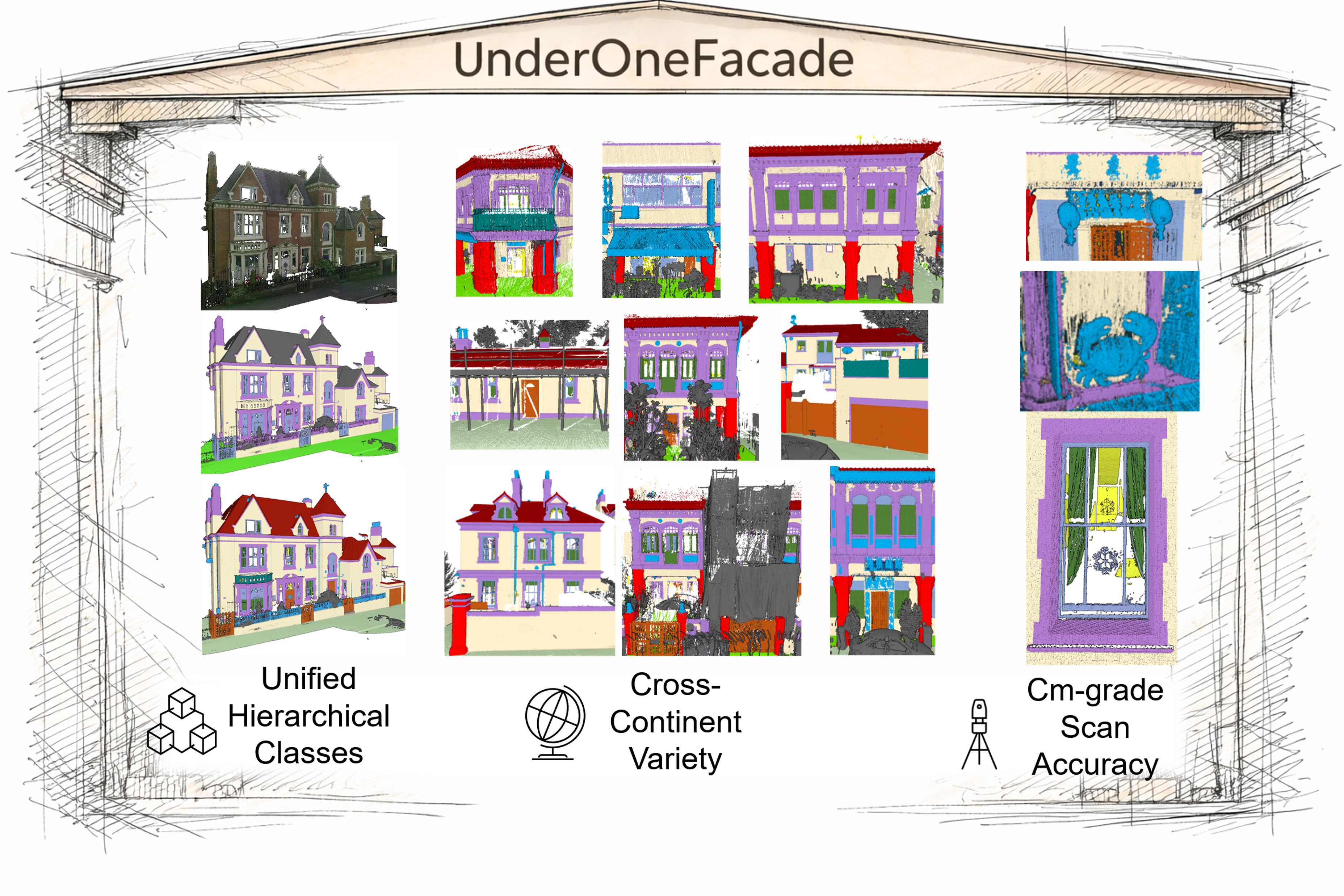}
  \caption{UnderOneFacade: We present a centimeter-accurate and cross-continental facade point clouds, with fine-grained semantic segmentation of architectural elements, and hierarchical facade taxonomy enabling multi-level evaluation (LoFG2–LoFG3), totaling 2.7B annotated points.}
  \label{fig:hierarchySemantics}
\end{figure*}

\begin{abstract}
Globally consistent semantic digital twins require centimeter-accurate and geographically transferable 3D facade segmentation. However, progress in facade parsing is limited by the lack of large-scale, standardized benchmarks for evaluating cross-domain and cross-country generalization. Existing datasets are geographically narrow, sensor-specific, semantically inconsistent, or insufficiently precise.
We introduce \textbf{UnderOneFacade}, the largest cross-continental 3D facade benchmark to date, comprising centimeter-accurate point clouds with hierarchical, harmonized, and architecturally grounded semantic labels totaling 2.7 billion annotated points. 
Through a systematic evaluation of representative point-, graph- and transformer-based architectures, we show that current methods struggle to recognize fine-grained architectural elements and degrade significantly across geographic regions, with the best models achieving only up to 33 IoU on the fine-grained LoFG3 benchmark.
By combining geometric precision with standardized semantics at unprecedented scale, UnderOneFacade establishes a rigorous benchmark for developing robust and transferable 3D segmentation models for and beyond facade understanding. 
The dataset, evaluation scripts, and pretrained models are available here: \url{https://jiangyuanwangyi.github.io/UnderOneFacade_official/}
\end{abstract}
\section{Introduction}
\label{sec:intro}
Facade semantic segmentation has been studied since the early days of computer vision \cite{szeliski2010computer}, owing to the structured yet intricate patterns present in building facades and their importance for urban scene understanding, large-scale mapping, and semantic digital twins. Accurate per-element labeling of facade components enables downstream applications such as architectural analysis, urban planning, energy modeling, and change detection \cite{biljeckiApplications3DCity2015,gehrung2022change,wysocki2023scan2lod3,Kolbe2021,matrone2020comparing}. Despite steady progress in 2D and 3D scene segmentation, facade understanding remains challenging due to its fine-grained structure, high intra-class variability, and strong dependence on precise geometric detail.

A key difficulty arises from the long-tail distribution of facade elements commonly observed in real-world data (e.g., ScanNet \cite{dai2017scannet}, A2D2 \cite{geyer2020a2d2}). While dominant classes such as walls are well represented, semantically critical architectural elements appear far less frequently, including doors, windows, and balconies. This imbalance mirrors challenges observed across indoor and outdoor semantic segmentation benchmarks \cite{dai2017scannet,yeshwanth2023scannet++,geyer2020a2d2,hackel2017semantic3d,Wysocki_2025_WACV} and significantly limits the effectiveness of data-driven approaches. As a result, models often produce biased predictions favoring dominant classes while failing to recognize smaller and structurally intricate facade elements even within the same architectural style \cite{archDatasetPaper,Wysocki_2025_WACV}.

These challenges are further exacerbated by the limited availability and diversity of annotated facade datasets. Existing benchmarks are typically limited to small-scale acquisitions covering individual buildings or homogeneous building blocks within narrow geographic regions (e.g., ZAHA \cite{Wysocki_2025_WACV}, ArCH \cite{archDatasetPaper}). Consequently, current models tend to overfit dataset- and country-specific architectural styles and struggle to generalize across cities or countries with different construction practices and design principles.

A seemingly straightforward solution is to leverage widely available street-level imagery to expand facade annotations. However, such data cannot provide the centimeter-level geometric accuracy of laser scanners required for precise architectural parsing. Facade segmentation is inherently accuracy-critical: even small geometric deviations can corrupt the delineation of fine structural elements and undermine semantic validity.

Beyond geometric fidelity, consistent semantic interpretation is equally important. Architecturally meaningful reasoning requires unified, standard-grounded class definitions rather than ad hoc, dataset-specific taxonomies. Without harmonized semantic structures transferable across cities and countries, benchmarking remains inconsistent and cross-domain evaluation unreliable.

The absence of large-scale 3D facade datasets that jointly ensure centimeter-grade geometric fidelity and standardized, architecturally grounded semantics therefore remains a core limitation of current benchmarks.

To address these limitations, we introduce \textbf{UnderOneFacade}, the largest cross-country 3D facade benchmark to date, comprising centimeter-accurate point clouds with unified semantic annotations grounded in architectural standards and totaling \textbf{2.7 billion annotated points}. The dataset spans multiple cities and architectural styles across different countries, enabling systematic evaluation of cross-country and cross-continent generalization. Our experiments show that even strong baseline architectures achieve only around \textbf{33 mIoU} on the most challenging LoFG3 benchmark, highlighting that fine-grained facade understanding remains far from solved.
Beyond facade parsing itself, the proposed benchmark also provides a controlled setting for studying broader problems in 3D scene understanding, including cross-domain generalization across architectural styles, long-tailed recognition of structural elements, and multimodal geometric–radiometric reasoning.
Our contributions are summarized as follows:
\begin{itemize}
    \item We introduce \textbf{UnderOneFacade}, the largest 3D facade segmentation benchmark with centimeter-level geometric accuracy and unified architectural semantics, comprising \textbf{2.7B annotated points}.
    \item We enable the first systematic evaluation of cross-country and cross-continent generalization in facade segmentation.
    \item We provide a comprehensive benchmark revealing the remaining challenges of fine-grained facade segmentation under realistic long-tailed distributions.
\end{itemize}
\section{Related Works}
\noindent\textbf{3D Semantic Segmentation Methods.}
Early approaches imposed regular structure on unorganized point clouds (e.g., voxels), enabling convolutional processing but introducing discretization artifacts and limiting scalability \cite{li2018pointcnn,dai20183dmv,riegler2017octnet,maturana2015voxnet}. 
More recent methods process raw points directly. Point-based networks such as PointNet and PointNet++ \cite{qi2017pointnet,qi2017pointnet++} introduced hierarchical aggregation to capture local geometric context, which is critical for thin and repetitive architectural structures. 
Subsequent work explored graph-based and transformer-based architectures to model local and global dependencies more effectively \cite{phan2018dgcnn,zhao2021point,wu2022point,lai2022stratified,wang2023octformer,yang2025swin3d}. Some approaches additionally incorporate hierarchical spatial partitions (e.g., octrees) to balance geometric precision and computational efficiency \cite{robert2023efficient,zhou2021adaptive}. 
Despite these advances, accurately segmenting fine-grained facade elements remains challenging due to long-tail class distributions and the limited availability of large-scale facade datasets.

\noindent\textbf{3D Urban Semantic Segmentation Datasets.}
Large-scale urban point cloud datasets have steadily expanded over the past decade, driven by advances in LiDAR sensing and 3D reconstruction. These benchmarks have enabled substantial progress in urban semantic segmentation and scene understanding. Representative datasets include Semantic3D \cite{hackel2017semantic3d}, Paris-Lille-3D \cite{roynard2018parisLille}, and Toronto-3D \cite{tan2020toronto}. 
However, these datasets primarily target coarse semantic categories such as buildings, roads, and vegetation. Fine-grained facade-level parsing is typically absent or treated marginally, limiting progress toward detailed architectural reasoning in 3D.
\begin{table*}[!htb]
\captionsetup{font=small}
\caption{Urban point cloud benchmark datasets for semantic segmentation. Columns indicate multi-national scope, multi-sensor acquisition, and whether hierarchical semantic classes are provided.}
\label{tab:bigTable}
\centering
\scriptsize
\setlength{\tabcolsep}{6pt}
\begin{adjustbox}{width=\textwidth,center}
\begin{tabular}{lccccccccc}
\toprule
\textbf{Name} & \textbf{Year} & \textbf{Sensor} & \textbf{Real / Synth} &
\textbf{\#Cls} & \textbf{Facade} & \textbf{\#Facade-Labeled} &
\textbf{Multi-} & \textbf{Multi-} & \textbf{Hierarchical} \\
& & & & & \textbf{classes?} & \textbf{points} &
\textbf{National?} & \textbf{Sensor?} & \textbf{classes?} \\
\midrule

Oakland 3D \cite{Munoz-2009-10227} & 2009 & MLS & real & 44 & $\thicksim$ & <0.1B & \redxmark & \redxmark & \redxmark \\
ETH PRS \cite{ethprs} & 2012 & TLS & real & 0 & \redxmark & \redxmark & \redxmark & \redxmark & \redxmark \\
Sydney Urban Objects \cite{SydneyDatasetde2013unsupervised} & 2013 & MLS & real & 26 & \redxmark & \redxmark & \redxmark & \redxmark & \redxmark \\
Paris-rue-Madame \cite{serna2014parisMadame} & 2014 & MLS & real & 27 & $\thicksim$ & <0.1B & \redxmark & \redxmark & $\thicksim$ \\
iQumulus \cite{vallet2015terramobilita} & 2015 & MLS & real & 8 & \redxmark & \redxmark & \redxmark & \redxmark & \redxmark \\
TUM-MLS-2016 \cite{zhu_tum-mls-2016_2020} & 2016 & MLS & real & 9 & \redxmark & \redxmark & \redxmark & \redxmark & \redxmark \\
semantic3D.net \cite{hackel2017semantic3d} & 2017 & TLS & real & 9 & \redxmark & \redxmark & \greencheck & \redxmark & \redxmark \\
Paris-Lille-3D \cite{roynard2018parisLille} & 2018 & MLS & real & 50 & \redxmark & \redxmark & \redxmark & \redxmark & \redxmark \\
SynthCity \cite{griffiths2019synthcity} & 2019 & MLS & synthetic & 9 & \redxmark & \redxmark & \redxmark & \redxmark & \redxmark \\
A2D2 \cite{geyer2020a2d2} & 2020 & MLS & real & 38 & \redxmark & \redxmark & \redxmark & \redxmark & \redxmark \\
ArCH \cite{matrone2020comparing} & 2020 & TLS/MLS/UAV & real & 10 & \greencheck & 0.1B & \redxmark & \greencheck & \redxmark \\
Toronto-3D \cite{tan2020toronto} & 2020 & MLS & real & 8 & \redxmark & \redxmark & \redxmark & \redxmark & \redxmark \\
Whu-TLS \cite{dong2020registration} & 2020 & TLS & real & 0 & \redxmark & \redxmark & \redxmark & \redxmark & \redxmark \\
BIMAGE \cite{BimageBlaser} & 2021 & MLS & real & 0 & \redxmark & \redxmark & \redxmark & \redxmark & \redxmark \\
KITTI-360 \cite{liao2021kitti} & 2021 & MLS & real & 19 & \redxmark & \redxmark & \redxmark & \redxmark & \redxmark \\
Paris-CARLA-3D \cite{deschaud2021pariscarla3d} & 2021 & MLS & real/synth & 23 & \redxmark & \redxmark & \greencheck & \redxmark & \redxmark \\
TUM-FAÇADE \cite{tumfacadePaper} & 2022 & MLS & real & 17 & \greencheck & 0.1B & \redxmark & \redxmark & \redxmark \\
HelixNet \cite{helixnet} & 2022 & MLS & real & 9 & \redxmark & \redxmark & \redxmark & \redxmark & \redxmark \\
SUD \cite{SUDdata_SiliviaGonzalez} & 2023 & MLS & real & 8 & \redxmark & \redxmark & \redxmark & \redxmark & \redxmark \\
ZAHA \cite{Wysocki_2025_WACV} & 2025 & MLS & real & 15 & \greencheck & 0.6B & \redxmark & \redxmark & \greencheck \\
Tuscany \cite{tuscanyData} & 2025 & MLS & real & 10 & \greencheck & <0.1B & \redxmark & \redxmark & \redxmark \\
TrueCity \cite{nguyentruecity} & 2025 & MLS & real/synth & 12 & \greencheck & 0.1B & \redxmark & \redxmark & \redxmark \\
City-Facade \cite{chen2026city} & 2026 & MLS & real & 9 & \greencheck & 0.2B & \redxmark & \redxmark & \redxmark \\

\midrule
\textbf{UnderOneFacade (Our)} & 2026 & TLS/MLS & real & 15 & \greencheck & 2.7 B & \greencheck & \greencheck & \greencheck \\
\bottomrule
\end{tabular}
\end{adjustbox}
\end{table*}

\noindent\textbf{Facade-Level 3D Datasets.}
Motivated by applications such as digital twins and automated building reconstruction \cite{wysocki2023scan2lod3,pantoja2022generating}, several datasets have begun to target facade-specific semantic segmentation with higher geometric fidelity and class granularity. Examples include ArCH \cite{matrone2020comparing}, TUM-FAÇADE \cite{tumfacadePaper}, and ZAHA \cite{Wysocki_2025_WACV}. 
Nevertheless, existing datasets remain limited in scale, architectural diversity, and geographic coverage. Their restricted size constrains deep model training and systematic evaluation. This bottleneck largely stems from the high cost of centimeter-accurate acquisition and the labor-intensive annotation of small, repetitive, and structurally complex facade elements.

\noindent\textbf{Semantic Consistency.}
Beyond dataset scale, facade benchmarks also exhibit substantial variability in class definitions, often relying on ad hoc or project-specific taxonomies. Such inconsistency impedes cross-dataset evaluation and reproducibility. 
The ZAHA dataset introduces the Level-of-Facade-Granularity (LoFG) concept, a hierarchical taxonomy grounded in architectural standards. While this represents an important step toward semantic harmonization, its realization within a single German city limits architectural variability and effectively validates the concept primarily on Central European typologies.
These limitations highlight the need for a large-scale, centimeter-accurate, cross-country 3D facade benchmark with standardized and architecturally grounded semantic labels. 

UnderOneFacade addresses this gap by providing the largest harmonized facade dataset to date, enabling systematic evaluation across countries and supporting the development of long-tail-robust and transferable 3D facade segmentation methods. In this sense, UnderOneFacade can be viewed as a cross-country extension of ZAHA: we inherit the LoFG taxonomy and the German subset from ZAHA, while contributing a $4.5\times$ increase in scale, multi-sensor acquisition, and a cross-country and cross-continent evaluation protocol.

\section{Dataset Design}

UnderOneFacade prioritizes highest possible geometric accuracy by deploying laser scanners with up to centimeter local accuracy and globally georeferenced centimeter acquisition, increasing architecture pattern richness by acquiring cross-continent data, multi-modal scanner setups to ensure scanner-related patterns variability but still ensure even cm-grade accuracy, and ensuring high-definition labels by manual and hierarchical annotations.

Three representative regions are selected to capture architectural diversity across continents. 
The European subsets (UK and Germany) share historical influences but differ in facade materials, ornamentation, and structural composition (Victorian vs Haussmann). 
In contrast, the Singapore subset represents Southeast Asian urban architecture, combining traditional Malay and Chinese elements with modern high-density construction. 
This selection introduces substantial stylistic variation and enables systematic evaluation of cross-regional generalization in facade segmentation.
\subsection{Laser Scans}

\sloppy

\noindent\textbf{UK.}
A static terrestrial laser scanning (TLS) campaign was conducted in The Park residential district in Nottingham, United Kingdom, using the Leica RTC360 \cite{leicaRTC360} scanner, which provides local accuracy at the centimeter-level and rapid full-dome capture. 
A total of 17 scanning stations were placed along the urban segment to ensure full facade coverage, reduce occlusions, and maintain sufficient overlap between neighboring scans.
To ensure consistent global alignment, the survey was fully georeferenced using a network of ground control points (GCPs) measured with GNSS and RTK corrections, providing centimeter-level positional accuracy in the national coordinate system. 
Individual scans were initially aligned using the scanner’s visual–inertial system and subsequently refined through global registration.
The final dataset captures dense facade geometries with fine architectural detail, including window frames, balconies, cornices, and recessed entrances. 
Compared to mobile mapping data, the static multi-station acquisition significantly increases point density on vertical structures while reducing occlusions.
Using the georeferenced data, individual buildings were extracted using OpenStreetMap (OSM) building footprints \cite{geofabrik}, expanded with a 5\,m buffer to account for footprint inaccuracies. 
The resulting UK subset contains 997M points across approximately 100 facades. 
No additional noise filtering was applied in order to preserve raw acquisition fidelity.

\noindent\textbf{Germany.}
The German subset is based on the ZAHA dataset \cite{Wysocki_2025_WACV}, which we integrate into the UnderOneFacade benchmark. 
ZAHA builds upon the publicly available TUM-MLS-2016 dataset \cite{zhu_tum-mls-2016_2020}, acquired in the Munich downtown area using the Mobile Distributed Situation Awareness (MODISSA) platform.
The platform is equipped with two Velodyne HDL-64E LiDAR sensors mounted at the front of the vehicle and two Velodyne VLP-16 sensors at the rear, providing dense multi-view urban coverage. 
Georeferencing is ensured through an inertial navigation system combined with RTK corrections from the German satellite positioning service (SAPOS) \cite{borgmann2018data,zhu_tum-mls-2016_2020}.
The dataset captures diverse architectural styles ranging from late 19th century historic buildings to modern urban developments, including residential, commercial, and cultural heritage structures. 
Following the same OSM-based extraction procedure as used for the UK subset, facade point clouds were clipped per building footprint with a 5\,m buffer.
The resulting German subset contains 601M points across approximately 66 facades. 
No additional noise removal was applied.

\noindent\textbf{Singapore.}
For the Singapore subset, data were collected using the Leica BLK360 \cite{leicaBLKarc} laser scanner, which captures high-density point clouds with centimeter-level accuracy. 
Scanning was performed with the scanner mounted to a backpack and from street-level positions along major urban corridors, carefully selected to cover a broad range of architectural styles from modern high-rise buildings to colonial-era structures.
Individual scans were registered using geometric feature alignment and subsequently referenced to geographic coordinates using OpenStreetMap (OSM) data. 
The registered point clouds were merged into a unified coordinate frame, producing a dense and geometrically consistent dataset suitable for facade-level analysis.
As with the other subsets, buildings were extracted using OSM footprints with a 5\,m buffer. 
The Singapore subset contains 1.117B points across approximately 200 facades. 
No additional noise filtering was applied.

\subsection{Hierarchical Semantic Classes Annotations}
\begin{figure}[ht]
  \centering
  \includegraphics[width=\textwidth]{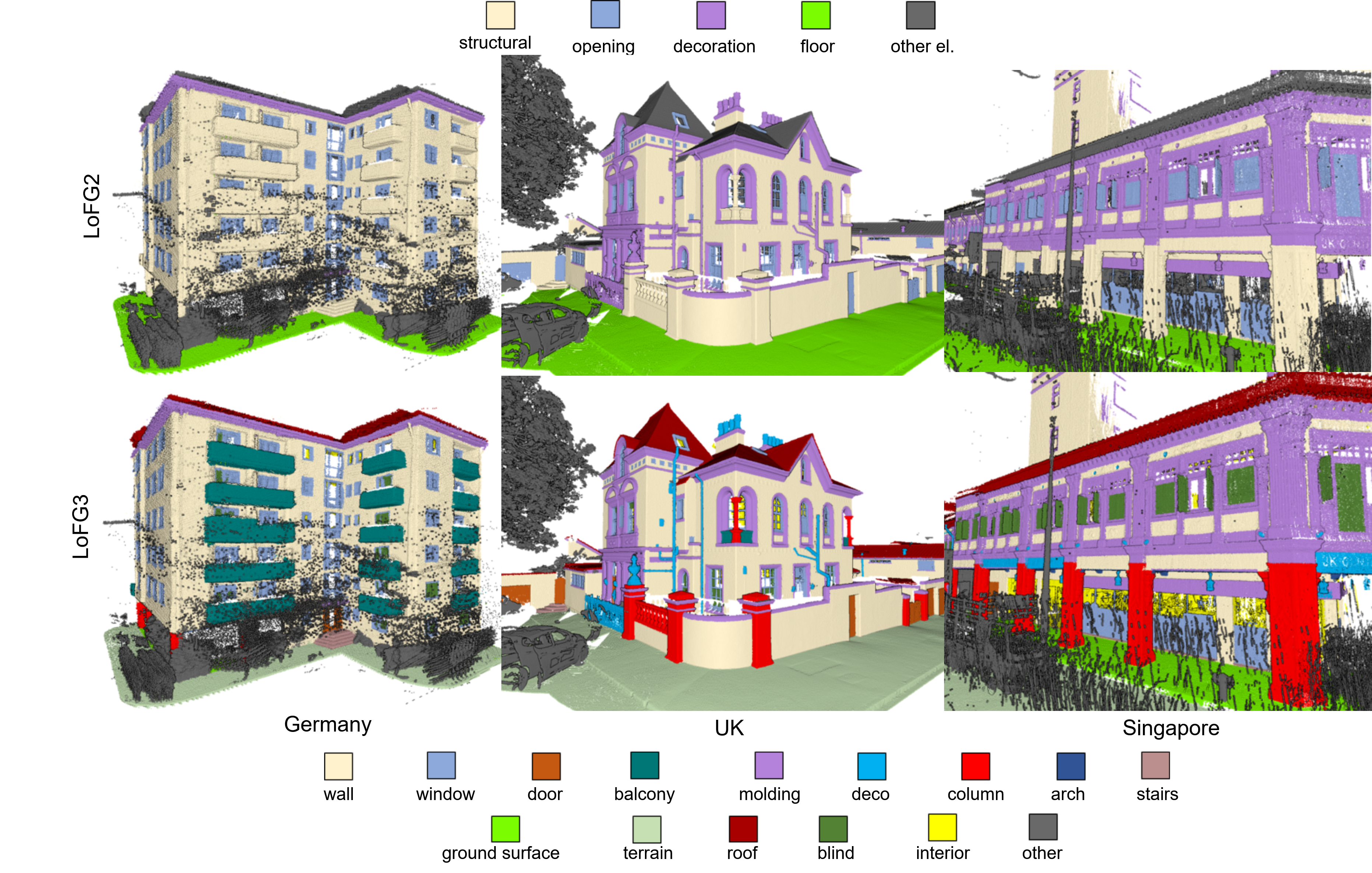}
  \label{fig:lofgHierarchy}
\end{figure}
\noindent\textbf{Hierarchical Classes.} High-fidelity 3D facade semantic segmentation requires both hierarchical class definitions and cross-country consistency. To address the lack of benchmarks with sufficient facade variability and standardized semantics, we adopt the Level of Facade Generalization (LoFG) concept \cite{Wysocki_2025_WACV} to unify semantic labels across diverse architectural contexts. LoFG defines three levels of detail (LoFG1, LoFG2, and LoFG3) allowing precise formulation of segmentation tasks, supporting transfer learning and label propagation, and enabling consistent evaluation across datasets and regions.

In this work, we cross-country-adopted LoFG to provide 15 detailed classes at LoFG3, aggregated into 5 coarser classes at LoFG2, and a single abstract class at LoFG1. The adaptation ensures that the taxonomy remains valid across countries with distinct architectural styles while adhering to international urban modeling and architectural standards such as CityGML, IFC, and the Art and Architecture Thesaurus (AAT) \cite{kutznerCityGMLNewFunctions2020,laakso2012ifc,matrone2020comparing}. LoFG3 includes fine-grained elements such as wall, balcony, molding, decorative elements, stairs, columns, arches, windows, blinds, doors, roof, ground, terrain, interior, and other (mainly noise). LoFG2 merges these into broader functional categories of structural (wall, columns, arch, balcony, stairs), opening (blinds, door, window), decoration (molding, deco), floor (ground surface, terrain), and other elements (roof, other, interior) classes, while LoFG1 abstracts the entire facade.

\noindent\textbf{Annotation Process.}
All annotations were performed manually by trained annotators with expertise in architecture and geospatial modeling, following the architectural standards described earlier. 
%
Labeling was carried out in professional large-scale point cloud software, Cyclone 3DR \cite{leicaCyclone}, enabling precise interaction with dense point clouds. 
Annotating a single facade required approximately 15 hours, resulting in roughly 5,500 hours of manual annotation for the full dataset of about 366 facades.
All annotations were further harmonized and cross-checked through annotator consensus, with a correction rate of around 10\%, occurring mainly in ambiguous categories (e.g., \textit{molding} vs.\ \textit{decoration}) as defined in our annotation guidelines (Suppl.\ Sec.~9).

\noindent\textbf{Data Splits.}
Each country subset is divided into training, validation, and test splits following a 70/20/10 ratio based on the total number of points. 
Splitting is performed per building, ensuring that complete facades belong to only one subset and that the splits are spatially disjoint.
It was also ensured to preserve both dominant and underrepresented semantic classes across all subsets. 
For reference, mean facade heights are 8.2/10.6/12.1\,m, while mean point densities are 11,483/28,713/33,199\,pts/m$^2$ before and 575/427/2,446\,pts/m$^2$ after voxelization (Singapore/UK/Germany, respectively).

\begin{figure}[t]
    \centering
    \begin{subfigure}[b]{0.48\textwidth}
        \centering
        \includegraphics[width=\textwidth]{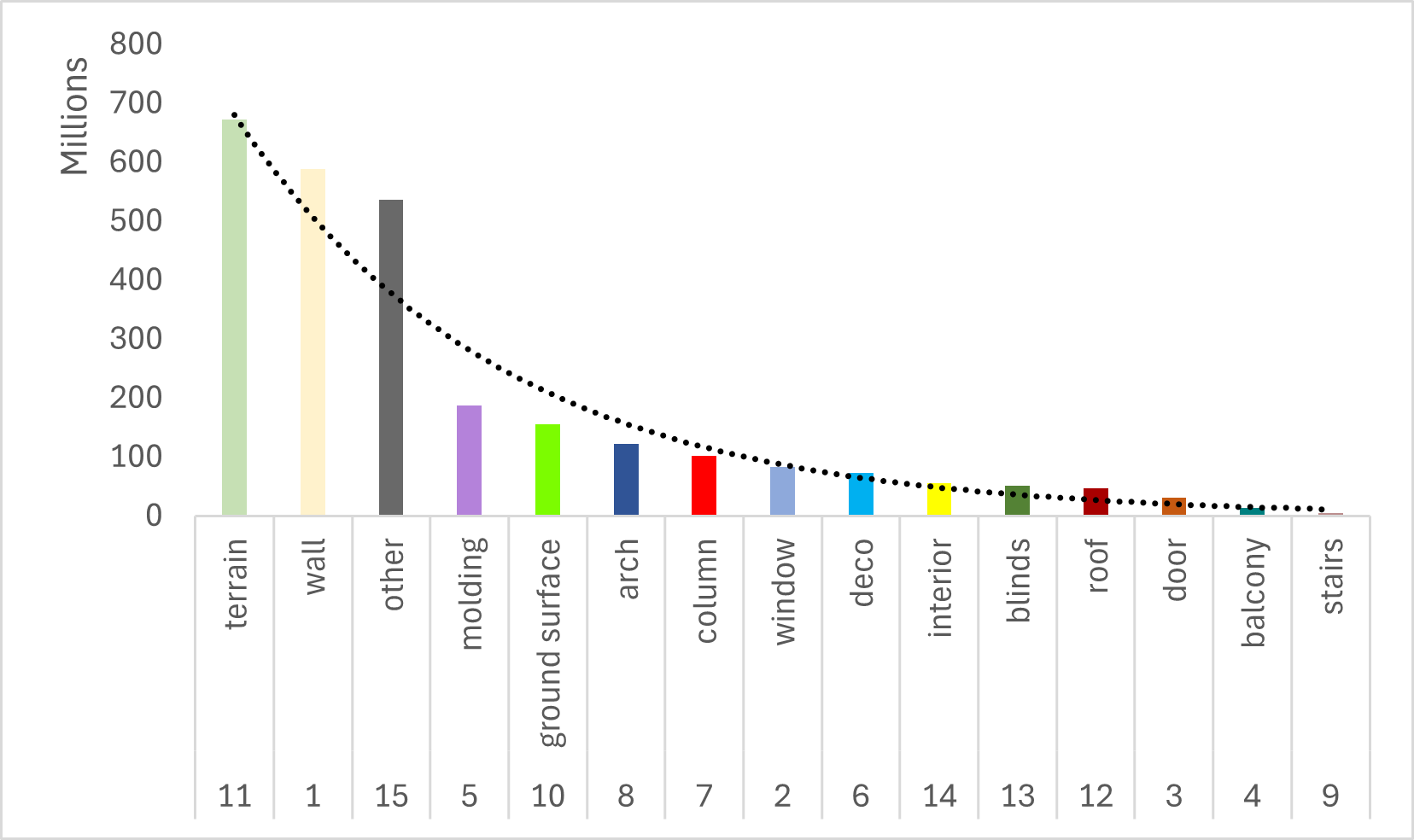}
        \caption{UnderOneFacade (our) with 2.7 BN points across countries.}
        \label{fig:sub1}
    \end{subfigure}
    \hfill
    \begin{subfigure}[b]{0.48\textwidth}
        \centering
        \includegraphics[width=\textwidth]{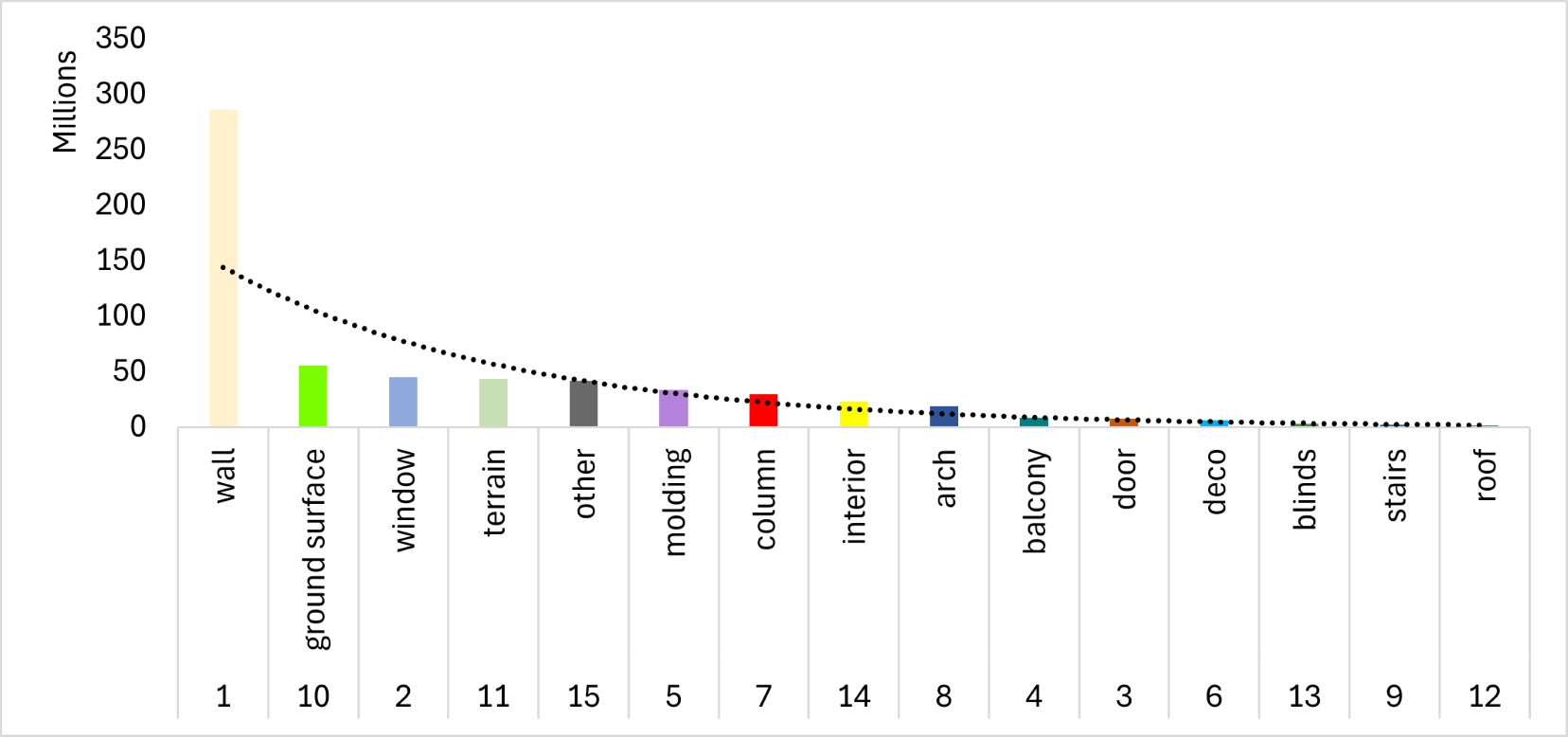}
        \caption{ZAHA \cite{Wysocki_2025_WACV} with 0.6 BN points in one country.}
        \label{fig:sub2}
    \end{subfigure}
    \caption{Comparison of per-point class distributions of UnderOneFacade (2.7BN points) and ZAHA (0.6 BN). Cross-country aggregation yields a substantially heavier-tailed and structurally more diverse distribution.}
    \label{fig:zaha_dist}
    \vspace{-1\baselineskip}
\end{figure}

\noindent\textbf{UnderOneFacade Benchmark Challenges}
UnderOneFacade is designed to evaluate facade segmentation along three
benchmark axes: (a) cross-country domain shifts, (b) hierarchical semantic
granularity, and (c) realistic long-tailed class distributions.
As shown in \cref{fig:zaha_dist}, aggregating multiple countries results in a
markedly heavier-tailed distribution compared to single-country datasets, opening new challenges.
By combining centimeter-accurate geometry, harmonized LoFG semantic labels, and architectural diversity across continents, the benchmark enables systematic evaluation of segmentation robustness beyond single-country datasets and can serve as a generic long-tail benchmark beyond facade segmentation.
These three principles define the core evaluation axes of UnderOneFacade. In the following section, we detail the experimental protocol designed to systematically assess architectural sensitivity, cross-domain generalization, and long-tail robustness across representative point cloud segmentation architectures.

\noindent\textbf{Release.} The dataset is released under an open data license (CC BY 4.0, Creative Commons Attribution), hosted on a cloud service (i.e., Google Drive \url{https://drive.google.com/drive/folders/1Yzz7PmyeK1qeOtkTFCfkbw7IEHXcMJo8?usp=sharing}) to allow unrestricted community use; the project page with additional details: \url{https://jiangyuanwangyi.github.io/UnderOneFacade_official/}.
No ethical approval was required for this study as it did not involve human or animal subjects.

\section{Experimental Setup}

\noindent\textbf{Baselines.}
We select representative baselines from the major families of point cloud networks to reflect different underlying working principles.
Namely, point-based: PointNet++ \cite{qi2017pointnet++}, KPConv \cite{thomas2019kpconv}; graph-based: DGCNN \cite{phan2018dgcnn}; and transformer-based approaches: OctFormer \cite{wang2023octformer} , Point Transformer (abbrv. as PTv1) \cite{zhao2021point}, and Point Transformer v3 (abbrv. as PTv3) \cite{wu2024point}. 
This selection allows us to systematically analyze how performance varies depending on architectural design and feature aggregation strategy.
The chosen methods are arguably among the most widely used models in the literature and serve as strong reference points across numerous point cloud benchmarks and methods \cite{tan2020toronto,Wysocki_2025_WACV,yeshwanth2023scannet++,geyer2020a2d2}. 
Moreover, they are prevalent in related tasks such as facade segmentation and large-scale urban scene understanding \cite{Wysocki_2025_WACV,wysocki2023scan2lod3,matrone2020comparing}, making them particularly suitable for comparison in our setting.
To isolate the geometric difficulty of facade segmentation,
the primary benchmark uses only xyz coordinates as input (radiometric features impact is analyzed separately in \Cref{tab:nottingham_rgbi_clean} and accompanying paragraph). 


\noindent\textbf{Hyperparameters.}
For a fair and controlled comparison, we evaluate all baseline methods under a unified training protocol inspired by ZAHA~\cite{Wysocki_2025_WACV} (e.g., a fixed 100 training epochs), rather than performing per-method hyperparameter tuning.
To analyze the impact of facade architecture and acquisition on various methods under identical budgets, we intentionally used unweighted cross-entropy. And to maintain experimental consistency, we train each method for the same number of epochs and apply identical data preprocessing and uniform point sampling strategies across all networks.
We also deploy standard metrics to analyze the performance on the two chosen semantic granularities: Overal Accuracy (OA), Precision (P), Recall (R), F1 score (F1), and Intersection over Union (IoU).
More implementation details in Suppl.
\section{Results and Discussion}
\begin{figure}[!htbp]
\vspace{-5mm}
  \centering
  \includegraphics[width=\textwidth]{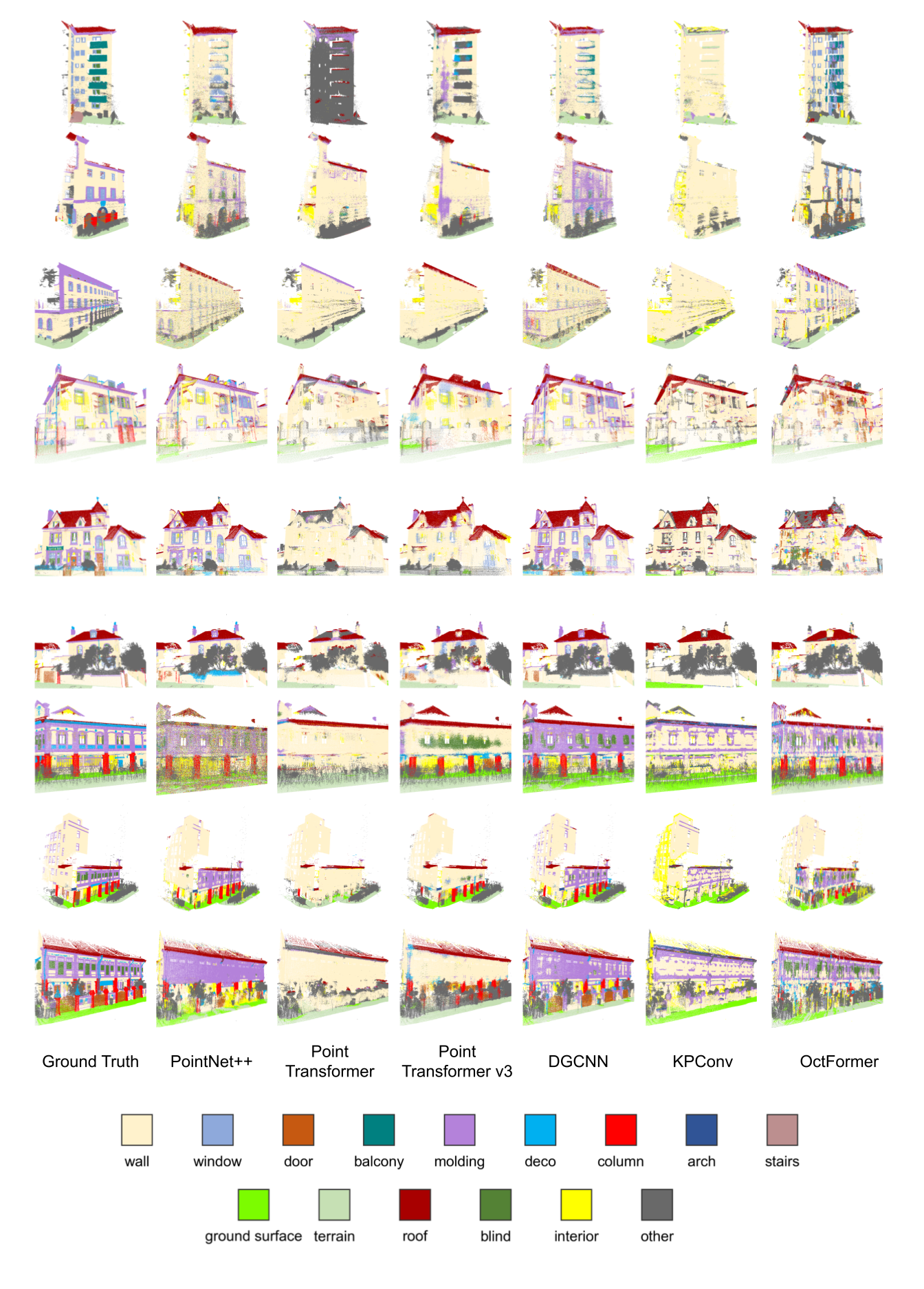}
  \caption{Qualitative facade segmentation results on UnderOneFacade on the LoFG3:
We visualize predictions of representative architectures across scenes from different countries.
Rows show example facades, while columns correspond to different segmentation models.
Despite correct segmentation of dominant structures such as \textit{walls} and \textit{roofs}, models struggle to consistently recognize fine-grained facade elements, including \textit{windows}, \textit{doors}, and decorative components.
These examples illustrate the challenges posed by long-tailed semantics and cross-country architectural variability.}
  \label{fig:lofg3}
\end{figure}

\noindent\textbf{Overall Performance on Hierarchical Semantic.}
Overall, the results show that fine-grained facade segmentation remains far from solved (\cref{tab:overallScores,fig:lofg3}).
\begin{table*}[h]
  \centering
  \vspace{-3mm}
  \scriptsize
  \setlength{\tabcolsep}{2pt}
  \renewcommand{\arraystretch}{0.85}
  \caption{Accuracy across outdoor long-tail datasets and architectures on the most challenging LoFG3. The Overall column reports models trained on the full UnderOneFacade dataset, while others solely within countries.}
  \begin{tabularx}{\textwidth}{@{}l|l|*{8}{>{\centering\arraybackslash}X}@{}}
    \toprule
    Family & Method &
    \multicolumn{2}{c}{UK} &
    \multicolumn{2}{c}{Singapore} &
    \multicolumn{2}{c}{Germany} &
    \multicolumn{2}{c}{Overall} \\
    
    \cmidrule(lr){3-4}
    \cmidrule(lr){5-6}
    \cmidrule(lr){7-8}
    \cmidrule(lr){9-10}

    & & OA & mIoU & OA & mIoU & OA & mIoU & OA & mIoU \\ 
    \midrule

    \multirow{2}{*}{Point}
    & PointNet++
      & 68.5 & 22.1
      & 59.6 & 29.8
      & 66.4 & 25.6
      & 60.9 & 29.7 \\

    & KPConv
      & 83.3 & 23.4
      & 41.5 & 11.2
      & 55.8 & 11.2 
      & 49.0 & 15.0 \\
      
    \midrule

    \multirow{1}{*}{Graph}
    & DGCNN
      & 77.8 & 28.3
      & 62.2 & 31.3
      & 71.1 & 33.4
      & 62.7 & 33.4 \\

    \midrule

    \multirow{3}{*}{Transformer}
    & PTv1
      & 79.0 & 31.0
      & 44.7 & 19.8
      & 75.0 & 41.6
      & 48.8 & 15.1 \\

    & PTv3
      & 79.0 & 26.5
      & 58.5 & 29.6
      & 56.8 & 15.7
      & 58.3 & 27.3 \\

    & OctFormer
      & 75.5 & 27.7
      & 61.5 & 34.6
      & 44.9 & 13.5
      & 57.9 & 27.3 \\

    \midrule
  \end{tabularx}
  \label{tab:overallScores}
  \vspace{-1\baselineskip}
\end{table*}
Even the strongest model reaches only about 33 mIoU on the combined dataset (DGCNN: 33.4), highlighting the difficulty of long-tailed facade semantics. Interestingly, the graph-based DGCNN consistently outperforms several more recent transformer-based architectures (e.g., PTv1: 15.1 and PTv3: 27.3 mIoU overall), suggesting that current transformer designs may not yet effectively capture the geometric structure of facade point clouds. While several architectures achieve high overall accuracy on individual datasets, their mIoU values remain limited, indicating that dominant classes such as \textit{wall} or \textit{roof} are segmented reliably whereas rare facade components remain challenging.
\begin{table*}[h]
\setlength\tabcolsep{4pt}
\scriptsize
\centering
\caption{Per-class F1 scores for complete UnderOneFacade (\%). Results are reported for two semantic granularity levels: LoFG3 (fine-grained) and LoFG2 (coarser). Methods are grouped by architectural family.}
\label{tab:overall_classes_combined}

\begin{tabular}{l|cc|c|ccc}
\hline
\textbf{Class} &
\multicolumn{2}{c|}{\textbf{Point-based}} &
\textbf{Graph-based} &
\multicolumn{3}{c}{\textbf{Transformer-based}} \\
\cline{2-7}
 & PointNet++ & KPConv & DGCNN & PT & PTv3 & OctFormer \\
\hline

\multicolumn{7}{c}{\textbf{LoFG3 (fine facade semantics)}} \\
\hline

wall & 70.5 & 61.2 & 71.7 & 60.5 & 65.2 & \textbf{71.9} \\
window & \textbf{37.2} & 0.0 & 36.9 & 14.7 & 5.1 & 28.1 \\
door & 13.9 & 0.0 & \textbf{26.7} & 0.2 & 16.4 & 10.3 \\
balcony & 19.3 & 0.0 & \textbf{46.7} & 0.0 & 6.4 & 43.4 \\
molding & 46.8 & 31.0 & \textbf{48.4} & 15.4 & 8.7 & 41.7 \\
deco & \textbf{10.0} & 0.0 & 9.2 & 1.0 & 9.2 & 9.6 \\
column & \textbf{45.7} & 0.0 & 43.7 & 1.1 & 43.3 & 29.5 \\
arch & 42.6 & 8.4 & 41.0 & 2.5 & \textbf{52.8} & 31.5 \\
stairs & 6.5 & 0.0 & \textbf{9.1} & 1.1 & 0.7 & 1.3 \\
ground surface & 54.1 & 48.3 & 54.8 & 20.2 & \textbf{57.9} & 51.6 \\
terrain & 70.4 & 44.6 & 73.8 & 64.6 & \textbf{75.0} & 67.7 \\
roof & 68.8 & 46.0 & \textbf{72.6} & 45.8 & 69.3 & 70.3 \\
blinds & 19.5 & 0.0 & \textbf{28.6} & 0.4 & 12.8 & 23.3 \\
interior & 61.9 & 35.0 & 69.9 & 53.5 & \textbf{74.6} & 53.6 \\
other & 70.7 & 63.0 & \textbf{73.1} & 55.0 & 69.5 & 63.4 \\

\hline
\multicolumn{7}{c}{\textbf{LoFG2 (coarse facade semantics)}} \\
\hline

floor & 93.3 & 66.7 & 72.2 & 85.8 & \textbf{94.0} & 69.7 \\
decoration & \textbf{45.1} & 0.0 & 45.1 & 4.1 & 30.4 & 36.0 \\
structural & \textbf{73.1} & 5.7 & 47.7 & 64.2 & 66.1 & 44.3 \\
opening & 44.3 & 83.8 & \textbf{92.1} & 20.1 & 25.3 & 87.8 \\
other & 75.4 & 66.6 & 75.5 & 67.0 & \textbf{75.6} & 71.7 \\

\hline
\end{tabular}
\vspace{-1\baselineskip}
\end{table*}
This observation is further confirmed by the per-class analysis in \Cref{tab:overall_classes_combined}. While aggregating classes into the coarser LoFG2 hierarchy slightly increases scores for dominant categories, it does not resolve the underlying segmentation difficulty. For example, the \textit{floor} class reaches high performance (e.g., 93.3 F1 for PointNet++), whereas structurally meaningful categories remain challenging. The \textit{decoration} class shows large variability across architectures (45.1 for PointNet++ vs.\ 4.1 for PT), and even the aggregated \textit{opening} class reveals strong architectural sensitivity, with transformer-based models performing substantially worse (PT: 20.1, PTv3: 25.3) than DGCNN (92.1). Similar patterns are observed in the fine-grained LoFG3 classes such as \textit{window}, \textit{door}, and \textit{molding}. Overall, semantic aggregation alone does not overcome the long-tailed distribution and geometric complexity of facade elements.

\noindent\textbf{Cross-Country Architectural Variability.}
\begin{table*}[t]
\setlength\tabcolsep{3pt}
\scriptsize
\centering
\caption{Per-class F1 comparison between UK and Singapore datasets (\%). 
Each cell shows \textit{UK / SG} and the difference $\Delta$ (SG$-$UK): red → Singapore harder, green → Singapore easier.}
\label{tab:SG_UK_delta}

\begin{tabular}{l|cc|c|ccc}
\hline
\textbf{Class} &
\multicolumn{2}{c|}{\textbf{Point-based}} &
\textbf{Graph-based} &
\multicolumn{3}{c}{\textbf{Transformer-based}} \\

& PointNet++ & KPConv & DGCNN & PT & PTv3 & OctFormer \\
\hline

\multicolumn{7}{c}{\textbf{LoFG3 (fine facade semantics)}} \\
\hline

wall &
\shortstack{72.5 / 54.4 \\ \gainn{-18.1}} &
\shortstack{81.1 / 49.5 \\ \gainn{-31.6}} &
\shortstack{79.5 / 58.0 \\ \gainn{-21.5}} &
\shortstack{83.7 / 43.8 \\ \gainn{-39.9}} &
\shortstack{81.7 / 50.8 \\ \gainn{-30.9}} &
\shortstack{81.1 / 64.4 \\ \gainn{-16.7}} \\

window &
\shortstack{13.7 / 3.1 \\ \gainn{-10.6}} &
\shortstack{0.0 / 0.0 \\ 0.0} &
\shortstack{27.9 / 2.6 \\ \gainn{-25.3}} &
\shortstack{23.1 / 5.2 \\ \gainn{-17.9}} &
\shortstack{10.9 / 0.0 \\ \gainn{-10.9}} &
\shortstack{19.6 / 9.1 \\ \gainn{-10.5}} \\

door &
\shortstack{5.7 / 0.6 \\ \gainn{-5.1}} &
\shortstack{0.0 / 0.0 \\ 0.0} &
\shortstack{10.3 / 0.3 \\ \gainn{-10.0}} &
\shortstack{21.7 / 0.0 \\ \gainn{-21.7}} &
\shortstack{0.0 / 0.0 \\ 0.0} &
\shortstack{21.8 / 17.6 \\ \gainn{-4.2}} \\

molding &
\shortstack{23.6 / 60.2 \\ \gainp{36.6}} &
\shortstack{31.7 / 18.5 \\ \gainn{-13.2}} &
\shortstack{40.7 / 63.3 \\ \gainp{22.6}} &
\shortstack{46.3 / 22.9 \\ \gainn{-23.4}} &
\shortstack{41.8 / 41.5 \\ \gainn{-0.3}} &
\shortstack{29.0 / 56.3 \\ \gainp{27.3}} \\

column &
\shortstack{2.8 / 59.6 \\ \gainp{56.8}} &
\shortstack{0.0 / 4.5 \\ \gainp{4.5}} &
\shortstack{14.1 / 62.9 \\ \gainp{48.8}} &
\shortstack{18.9 / 30.7 \\ \gainp{11.8}} &
\shortstack{0.0 / 62.1 \\ \gainp{62.1}} &
\shortstack{0.0 / 65.4 \\ \gainp{65.4}} \\

arch &
\shortstack{0.0 / 62.0 \\ \gainp{62.0}} &
\shortstack{0.0 / 23.8 \\ \gainp{23.8}} &
\shortstack{0.0 / 67.3 \\ \gainp{67.3}} &
\shortstack{0.0 / 57.6 \\ \gainp{57.6}} &
\shortstack{0.0 / 76.2 \\ \gainp{76.2}} &
\shortstack{0.0 / 80.0 \\ \gainp{80.0}} \\

terrain &
\shortstack{95.4 / 62.4 \\ \gainn{-33.0}} &
\shortstack{97.5 / 55.7 \\ \gainn{-41.8}} &
\shortstack{98.2 / 64.0 \\ \gainn{-34.2}} &
\shortstack{97.6 / 57.9 \\ \gainn{-39.7}} &
\shortstack{97.2 / 57.5 \\ \gainn{-39.7}} &
\shortstack{96.3 / 70.4 \\ \gainn{-25.9}} \\

roof &
\shortstack{77.3 / 75.2 \\ \gainn{-2.1}} &
\shortstack{87.8 / 7.6 \\ \gainn{-80.2}} &
\shortstack{84.7 / 80.4 \\ \gainn{-4.3}} &
\shortstack{87.0 / 64.4 \\ \gainn{-22.6}} &
\shortstack{88.7 / 83.7 \\ \gainn{-5.0}} &
\shortstack{87.2 / 80.5 \\ \gainn{-6.7}} \\

interior &
\shortstack{23.2 / 40.0 \\ \gainp{16.8}} &
\shortstack{0.0 / 0.8 \\ \gainp{0.8}} &
\shortstack{49.7 / 36.6 \\ \gainn{-13.1}} &
\shortstack{55.4 / 62.1 \\ \gainp{6.7}} &
\shortstack{47.7 / 69.6 \\ \gainp{21.9}} &
\shortstack{35.0 / 50.9 \\ \gainp{15.9}} \\

\hline
\multicolumn{7}{c}{\textbf{LoFG2 (coarse facade semantics)}} \\
\hline

floor &
\shortstack{94.9 / 94.8 \\ \gainn{-0.1}}&
\shortstack{97.7 / 93.0 \\ \gainn{-4.7}} &
\shortstack{97.6 / 97.9 \\ \gainp{0.3}} &
\shortstack{97.7 / 88.5 \\ \gainn{-9.2}} &
\shortstack{97.5 / 96.1 \\ \gainn{-1.4}} &
\shortstack{95.9 / 97.0 \\ \gainp{1.1}} \\

decoration &
\shortstack{21.5 / 60.5 \\ \gainp{39.0}} &
\shortstack{15.3 / 30.2 \\ \gainp{14.9}} &
\shortstack{40.6 / 51.1 \\ \gainp{10.5}} &
\shortstack{42.1 / 36.6 \\ \gainn{-5.5}} &
\shortstack{36.9 / 40.1 \\ \gainp{3.2}} &
\shortstack{34.0 / 49.4 \\ \gainp{15.4}} \\

structural &
\shortstack{71.9 / 61.7 \\ \gainn{-10.2}}&
\shortstack{83.0 / 47.7 \\ \gainn{-35.3}} &
\shortstack{79.9 / 66.3 \\ \gainn{-13.6}} &
\shortstack{82.1 / 52.4 \\ \gainn{-29.7}} &
\shortstack{82.0 / 60.3 \\ \gainn{-21.7}} &
\shortstack{79.8 / 65.5 \\ \gainn{-14.3}} \\

opening &
\shortstack{24.8 / 23.8 \\ \gainn{-1.0}}&
\shortstack{0.0 / 0.0 \\ 0.0} &
\shortstack{28.0 / 20.2 \\ \gainn{-7.8}} &
\shortstack{36.5 / 6.8 \\ \gainn{-29.7}} &
\shortstack{33.8 / 17.2 \\ \gainn{-16.6}} &
\shortstack{20.3 / 35.6 \\ \gainp{15.3}} \\

other &
\shortstack{82.9 / 72.8 \\ \gainn{-10.1}}&
\shortstack{89.6 / 66.6 \\ \gainn{-23.0}} &
\shortstack{88.1 / 67.5 \\ \gainn{-20.6}} &
\shortstack{88.9 / 55.5 \\ \gainn{-33.4}} &
\shortstack{90.0 / 78.5 \\ \gainn{-11.5}} &
\shortstack{86.5 / 79.8 \\ \gainn{-6.7}} \\

\hline
\end{tabular}
\vspace{-1\baselineskip}
\end{table*}
We analyze how architectural differences across countries affect facade segmentation, where \Cref{tab:SG_UK_delta} provides class-level comparisons between the most distinct UK and Singapore subsets.
At the fine-grained LoFG3 level, several classes exhibit substantial performance drops in Singapore, particularly \textit{wall} (e.g., PT: 83.7$\rightarrow$43.8, $-39.9$) and \textit{terrain} (97.6$\rightarrow$57.9, $-39.7$), indicating a clear architectural domain shift between European and Southeast Asian scenes. Similarly, detailed facade elements such as \textit{window} degrade strongly (DGCNN: 27.9$\rightarrow$2.6). Conversely, elements such as \textit{column} (2.8$\rightarrow$59.6) and \textit{arch} (0.0$\rightarrow$62.0) appear predominantly in Singapore, resulting in large positive gains.
At the coarser LoFG2 level, only \textit{floor} remains comparatively stable across countries, while \textit{structural} and \textit{other} drop for all architectures. Moreover, the \textit{opening} class degrades for most architectures (e.g., PT: 36.5$\rightarrow$6.8, $-29.7$; DGCNN: 28.0$\rightarrow$20.2), highlighting persistent difficulties in intricate facade openings of Southeast Asian buildings compared to the repetitive patterns of UK Victorian architecture.


\noindent\textbf{Impact of Radiometric Features.}
\begin{table*}[h]
\centering
\scriptsize
\setlength{\tabcolsep}{4pt}
\caption{Impact of radiometric features (RGBI) on the UK LoFG3 test set. 
$\Delta$ indicates the performance change when adding RGBI features ($+\mathrm{rgbi}-xyz$).}
\label{tab:nottingham_rgbi_clean}

\begin{tabular}{l|ccc|ccc|ccc}
\hline
 & \multicolumn{3}{c|}{\textbf{PointNet++}} 
 & \multicolumn{3}{c|}{\textbf{PT}} 
 & \multicolumn{3}{c}{\textbf{DGCNN}} \\

\cline{2-10}
\textbf{Metric/Class} 
& xyz & +rgbi & $\Delta$
& xyz & +rgbi & $\Delta$
& xyz & +rgbi & $\Delta$ \\

\hline
OA
& 68.5 & 73.9 & \gainp{5.4}
& 79.0 & 83.9 & \gainp{4.9}
& 77.8 & 76.5 & \gainn{1.3} \\

$\mu$P
& 28.9 & 39.0 & \gainp{10.1}
& 40.0 & 49.7 & \gainp{9.7}
& 40.5 & 43.1 & \gainp{2.6} \\

$\mu$R
& 27.8 & 37.7 & \gainp{9.9}
& 38.1 & 47.2 & \gainp{9.1}
& 33.4 & 37.6 & \gainp{4.2} \\

$\mu$F1
& 27.9 & 38.0 & \gainp{10.1}
& 38.5 & 47.9 & \gainp{9.4}
& 34.9 & 38.4 & \gainp{3.5} \\

$\mu$IoU
& 22.1 & 29.4 & \gainp{7.3}
& 31.0 & 39.1 & \gainp{8.1}
& 28.3 & 30.2 & \gainp{1.9} \\

\hline

wall
& 72.5 & 76.2 & \gainp{3.7}
& 83.7 & 86.1 & \gainp{2.4}
& 79.5 & 78.9 & \gainn{0.6} \\

window
& 13.7 & 48.8 & \gainp{35.1}
& 23.1 & 54.5 & \gainp{31.4}
& 27.9 & 38.2 & \gainp{10.3} \\

door
& 5.7 & 30.2 & \gainp{24.5}
& 21.7 & 44.8 & \gainp{23.1}
& 10.3 & 43.0 & \gainp{32.7} \\

balcony
& 0.0 & 0.0 & 0.0
& 0.0 & 0.0 & 0.0
& 0.0 & 0.0 & 0.0 \\

molding
& 23.6 & 41.5 & \gainp{17.9}
& 46.3 & 60.9 & \gainp{14.6}
& 40.7 & 47.1 & \gainp{6.4} \\

deco
& 12.1 & 18.1 & \gainp{6.0}
& 31.6 & 43.6 & \gainp{12.0}
& 20.5 & 34.2 & \gainp{13.7} \\

column
& 2.8 & 15.4 & \gainp{12.6}
& 18.9 & 33.9 & \gainp{15.0}
& 14.1 & 24.1 & \gainp{10.0} \\

arch
& 0.0 & 0.0 & 0.0
& 0.0 & 0.0 & 0.0
& 0.0 & 0.0 & 0.0 \\

stairs
& 0.0 & 0.0 & 0.0
& 0.0 & 0.0 & 0.0
& 0.0 & 0.0 & 0.0 \\

ground surface
& 0.0 & 0.0 & 0.0
& 0.0 & 0.0 & 0.0
& 0.0 & 0.0 & 0.0 \\

terrain
& 95.4 & 89.1 & \gainn{6.3}
& 97.6 & 97.8 & \gainp{0.2}
& 98.2 & 96.6 & \gainn{1.6} \\

roof
& 77.3 & 82.5 & \gainp{5.2}
& 87.0 & 92.7 & \gainp{5.7}
& 84.7 & 85.5 & \gainp{0.8} \\

blinds
& 8.3 & 17.0 & \gainp{8.7}
& 19.3 & 36.9 & \gainp{17.6}
& 6.0 & 0.3 & \gainn{5.7} \\

interior
& 23.2 & 64.4 & \gainp{41.2}
& 55.4 & 74.8 & \gainp{19.4}
& 49.7 & 39.8 & \gainn{9.9} \\

other
& 83.4 & 87.1 & \gainp{3.7}
& 93.1 & 93.2 & \gainp{0.1}
& 92.3 & 87.8 & \gainn{4.5} \\

\hline
\end{tabular}
\end{table*}
We analyze the impact of radiometric information by comparing models trained with geometric features only (xyz) against models using additional radiometric attributes, including projected RGB (rgb) values and laser intensity (i). 
\Cref{tab:nottingham_rgbi_clean} shows that incorporating radiometric features can improve performance, increasing mIoU by +7.3 for PointNet++ and +8.1 for PT. The improvements are most pronounced for facade elements with weak geometric signatures, such as \textit{window}, \textit{door}, and \textit{molding}, where gains exceed +30 F1 for windows and +24 F1 for doors, likely due to consistent color patterns across samples (e.g., white or brown doors and window frames). In contrast, large planar structures such as \textit{terrain} remain largely unaffected.
However, radiometric cues are not universally beneficial. For several classes, adding RGBI features leads to performance degradation, suggesting that models may learn dataset-specific radiometric shortcuts rather than robust geometric representations (e.g., OA -1.3 for DGCNN).
Overall, these results highlight that radiometric information can complement geometric features for small facade elements, but may also introduce appearance biases that limit robustness. Additional results are provided in the Suppl.



\noindent\textbf{Cross-Country Generalization and Distribution Shift.}
We evaluate cross-continental generalization under a zero-shot transfer protocol. Models are trained on one continental subset and evaluated directly on the other without any target-domain adaptation. Two transfer directions are considered: Europe $\rightarrow$ Asia (training on UK + ZAHA, testing on Singapore) and Asia $\rightarrow$ Europe. Both training sets contain comparable data volumes of roughly 1.2B points for Asia vs 1.5B points for Europe to ensure that observed differences reflect distributional rather than scale effects. Results are reported at both LoFG2 and LoFG3 levels.

\Cref{tab:cross_continent_xyz} reveals a clear asymmetry in transfer performance. At LoFG2, Asia$\rightarrow$Europe achieves a higher peak OA (65.8\%) than Europe$\rightarrow$Asia (60.9\%), while mIoU remains similar (40.5\% vs.\ 40.8\%). The asymmetry becomes more pronounced at LoFG3, where Asia$\rightarrow$Europe consistently outperforms Europe$\rightarrow$Asia across all methods, reaching 51.9\% OA and 23.6 mIoU compared to 42.6\% OA and 14.8 mIoU.
\begin{table*}[h]
    \centering
    \scriptsize
    \setlength{\tabcolsep}{3pt}
    \renewcommand{\arraystretch}{1.05}
    \caption{Cross-continental generalization (zero-shot) on LoFG2 and LoFG3 using xyz coordinates as input features (Europe: UK + ZAHA; Asia: Singapore), where a clear diagonal trend is distinguishable.}
    \begin{tabular}{@{}l l c c c c c c c c@{}}
    \toprule
    \multicolumn{2}{c}{} &
    \multicolumn{2}{c}{\shortstack{Europe $\rightarrow$ Asia\\(LoFG2)}} &
    \multicolumn{2}{c}{\shortstack{Europe $\rightarrow$ Asia\\(LoFG3)}} &
    \multicolumn{2}{c}{\shortstack{Asia $\rightarrow$ Europe\\(LoFG2)}} &
    \multicolumn{2}{c}{\shortstack{Asia $\rightarrow$ Europe\\(LoFG3)}} \\
    \cmidrule(lr){3-4}\cmidrule(lr){5-6}\cmidrule(lr){7-8}\cmidrule(lr){9-10}
    Family & Method &
    OA & mIoU &
    OA & mIoU &
    OA & mIoU &
    OA & mIoU \\
    \midrule
    \multirow{2}{*}{Point-based}
    & PointNet++ &
    56.4 & 33.9 &
    40.5 & 14.0 &
    \textbf{65.8} & 38.8 &
    \textbf{51.9} & \textbf{23.6} \\
    & KPConv &
    48.6 & 25.7 &
    32.1 & 6.6 &
    48.9 & 28.8 &
    33.9 & 9.4 \\
    \midrule
    Graph-based
    & DGCNN &
    56.4 & 38.3 &
    \textbf{42.6} & \textbf{14.8} &
    63.9 & \textbf{40.5} &
    44.8 & 17.4 \\
    \midrule
    \multirow{3}{*}{Transformer-based}
    & OctFormer &
    59.2 & 37.0 &
    39.0 & 11.5 &
    62.3 & 40.0 &
    50.5 & 16.2 \\
    & PTv1 &
    55.1 & 34.2 &
    33.8 & 14.0 &
    58.1 & 38.1 &
    41.9 & 20.4 \\
    & PTv3 &
    \textbf{60.9} & \textbf{40.8} &
    41.5 & 11.0 &
    53.4 & 34.2 &
    49.0 & 15.7 \\
    \bottomrule
    \end{tabular}
    \label{tab:cross_continent_xyz}
    \vspace{-1\baselineskip}
\end{table*}
Class distribution analysis provides a data-grounded explanation for this effect. The European training set is highly concentrated in dominant classes such as \textit{wall} and \textit{other}, which together account for 58.3\% of all points, while the Singapore set shows more balanced coverage across facade categories. Consequently, the Asian training data provides stronger supervision for rare classes, which improves transfer performance at the fine-grained LoFG3 level.

These results highlight that balanced semantic coverage in the source domain plays a key role in cross-continental generalization and underline the importance of geographically diverse datasets for robust facade understanding.
\subsection{Limitations and Future Work}
UnderOneFacade combines data from multiple sensing configurations, including vehicle-mounted mobile laser scanners, backpack systems, and static terrestrial scanners. 
While this increases acquisition diversity and realism, it makes it difficult to isolate the influence of individual sensor modalities without controlled acquisitions of the same environment.
We acknowledge that the observed cross-country and cross-continental domain shifts likely arise from a combination of factors, including sensor type, point density, acquisition geometry, and architectural style, rather than architecture alone. UnderOneFacade intentionally reflects these entangled real-world conditions through its per-country (sensor-homogeneous) and overall (sensor-heterogeneous) splits. We provide partial disentanglement via (i) uniform 5\,cm training-voxelization to reduce density differences across subsets, and (ii) the radiometric feature analysis in \Cref{tab:nottingham_rgbi_clean} and Suppl., which isolates radiometric shifts between the UK and Singapore subsets.
Furthermore, the dataset preserves the naturally occurring long-tailed distribution of facade elements. 
Although this reflects realistic urban scenes, it results in limited training samples for some rare classes. 
The current benchmark covers three cities, which, despite their architectural diversity, cannot fully represent the global variability of all facade structures. 
Nevertheless, the proposed hierarchical taxonomy and annotation protocol are designed to be extensible, enabling future integration of additional cities and sensing setups into the benchmark.
\section{Conclusion}
We present \textbf{UnderOneFacade}, the largest cross-country 3D facade segmentation benchmark to date, comprising centimeter-accurate point clouds with hierarchical semantic annotations totaling 2.7B labeled points. The dataset combines multiple sensing modalities and architectural styles across countries, producing realistic long-tailed distributions and substantial geometric diversity.
Our experiments show that the benchmark is highly challenging for current methods. Notably, transformer-based models do not consistently outperform earlier architectures, with DGCNN achieving the best overall performance, while still the highest IoU reaches only 33\%. Radiometric features offer limited benefit and can even degrade performance, indicating reliance on dataset-specific appearance shortcuts.
Cross-country evaluation further reveals strong architectural domain shifts, with several models degrading by more than 30 F1 between European and Asian scenes. These results highlight the limited robustness of existing models to architectural variability and long-tailed facade semantics.
We believe UnderOneFacade provides an important step toward more robust facade understanding and high definition urban digital twins and generic large-scale semantic segmentation challenges.



\section*{Acknowledgements}
UnderOneFacade is the result of a collaborative effort spanning several years, from approximately 2023 to 2026, building upon the earlier development of the ZAHA dataset. We gratefully acknowledge the support of the TUM Global Incentive Fund, which enabled a research stay at the National University of Singapore (NUS) and the acquisition of the Singapore component of the dataset.

We sincerely thank the researchers of the Urban Analytics Lab at NUS for their support during the Singapore data acquisition campaign, and acknowledge the ZAHA team and its annotators for providing the hierarchical foundation that inspired and enabled this work. We are especially grateful to Clara Väth for her meticulous annotation of the Singapore and UK subsets.

Finally, we thank our collaborators at the University of Nottingham for their hospitality and support during the early stages of this project, and the TUM Graduate School which made the visit in Nottingham possible.

This benchmark represents the combined efforts of many researchers, annotators, and collaborators across multiple institutions and countries. We are delighted to see this long-term collaborative effort culminate in the presentation of UnderOneFacade at ECCV 2026.

%
%
\bibliographystyle{splncs04}
\bibliography{main}
\end{document}